\documentclass{article}
\usepackage[top=4cm, bottom=4cm, left=4.1cm, right=4.1cm, includefoot]{geometry}
\usepackage{cite}
\usepackage{amsmath,amssymb,amsfonts}
\usepackage{algorithmic}
\usepackage{algorithm}
\usepackage{graphicx}
\usepackage{textcomp}
\usepackage{tabularx,booktabs}
\usepackage{enumitem}
\newcommand{\ra}[1]{\renewcommand{\arraystretch}{#1}}

\usepackage[numbers]{natbib}

\usepackage{hyperref}

\newcommand{\argmin}{\mathop{\mathrm{arg~min}}\limits}

\def\BibTeX{{\rm B\kern-.05em{\sc i\kern-.025em b}\kern-.08em
    T\kern-.1667em\lower.7ex\hbox{E}\kern-.125emX}}
\begin{document}

\title{Evolving Architectures with Gradient Misalignment toward Low Adversarial Transferability}
\author{Kevin Richard G. Operiano\footnote{Electrical Engineering Department, Chulalongkorn University, Bangkok 10330, Thailand},
Wanchalerm Pora\footnotemark[1],\\ Hitoshi Iba\footnote{Information Science and Technology, University of Tokyo, Tokyo 113-8654, Japan (e-mail: iba@iba.t.u-tokyo.ac.jp)}, and Hiroshi Kera\footnote{Graduate School of Engineering, Chiba University, Chiba 263-8522, Japan (e-mail: kera.hiroshi@gmail.com)}}

\maketitle
\begin{abstract}
Deep neural network image classifiers are known to be susceptible not only to adversarial examples created for them but even those created for others. This phenomenon poses a potential security risk in various black-box systems relying on image classifiers. The reason behind such transferability of adversarial examples is not yet fully understood and many studies have proposed training methods to obtain classifiers with low transferability. In this study, we address this problem from a novel perspective through investigating the contribution of the network architecture to transferability. Specifically, we propose an architecture searching framework that employs neuroevolution to evolve network architectures and the gradient misalignment loss to encourage networks to converge into dissimilar functions after training. 
Our experiments show that the proposed framework successfully discovers architectures that reduce transferability from four standard networks including ResNet and VGG, while maintaining a good accuracy on unperturbed images. In addition, the evolved networks trained with gradient misalignment exhibit significantly lower transferability compared to standard networks trained with gradient misalignment, which indicates that the network architecture plays an important role in reducing transferability.
This study demonstrates that designing or exploring proper network architectures is a promising approach to tackle the transferability issue and train adversarially robust image classifiers.
\end{abstract}

\section{Introduction}
\label{sec:introduction}

\begin{figure*}[t]
\centerline{\includegraphics[width=\textwidth]{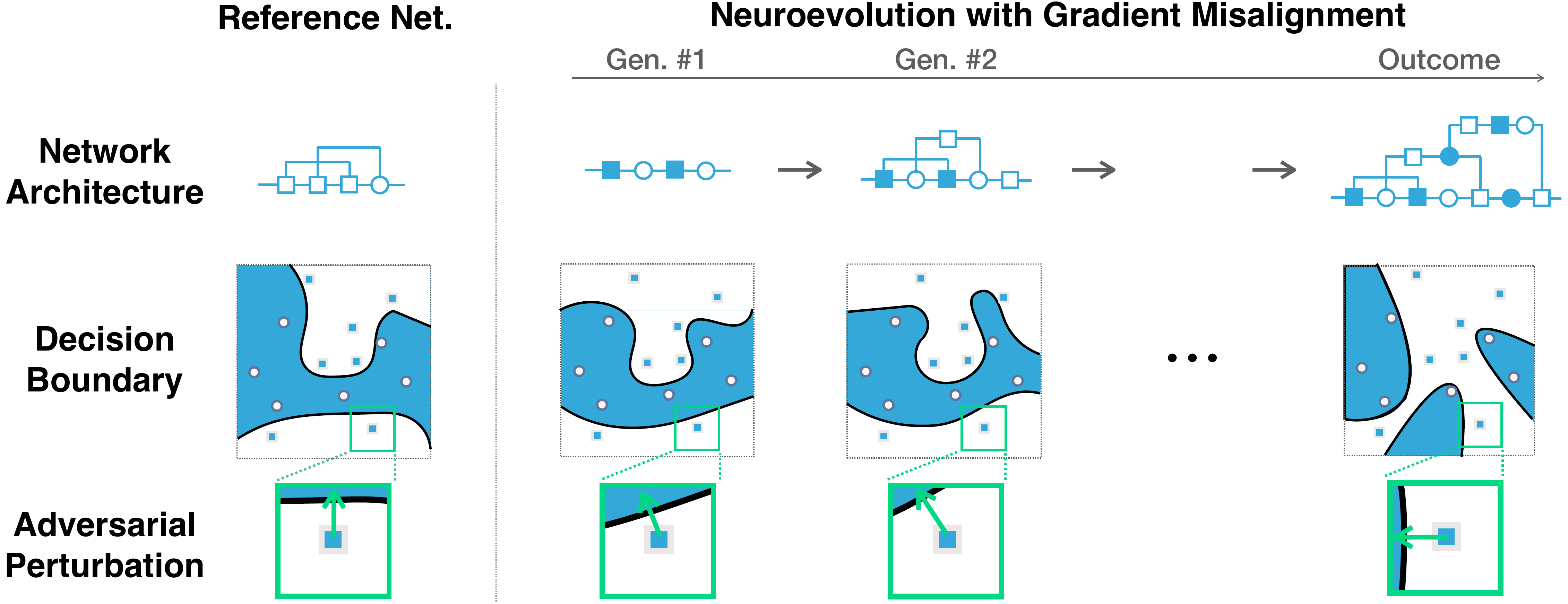}}
\caption{Overview of the proposed framework in a binary classification task. (Left pane) the reference network properly classifies the samples into two classes (circles and squares). (Right pane) Our framework uses neuroevolution to evolve network architectures (shown only one network per generation for illustration) in order to discover a network architecture by which one can train a classifier such that (i) it attains a high classification accuracy as the reference network and (ii) it has a drastically different direction of the adversarial perturbation (i.e., low transferability). The middle row shows that although the shape of decision boundaries are different across networks, the classification of samples are identical, which satisfies (i). The bottom row shows that through the course of evolution, the adversarial perturbation on the sample of interest changes, making it difficult to attack all the networks (in particular, the reference network and final outcome network) simultaneously, which satisfies (ii). }
\label{fig:overview}
\end{figure*}

Transferability of adversarial examples is a phenomenon where images that are slightly perturbed to fool an image classifier known as adversarial examples, can also fool other image classifiers whose parameters, training hyperparameters, and even architectures are different from those of the originally targeted classifier~\cite{goodfellow2014explaining, biggio2013evasion, dong2018boosting, liu2016delving, moosavi2017universal, papernot2017practical, papernot2016transferability, szegedy2013intriguing, tramer2017space, demontis2019adversarial}. The transferability of adversarial examples poses serious threats to many systems based on image classification (e.g., pedestrian detection in the autonomous driving) because even when attackers do not have a direct access to these systems, they may fool the systems by adversarial examples that are generated using other whitebox systems. Therefore, training an image classifier that achieves high classification accuracy on unperturbed images\footnote{Hereinafter, the classification accuracy on unperturbed images and adversarial examples are referred to as \textit{clean accuracy} and \textit{adversarial accuracy}, respectively.} and is also robust against adversarial examples generated from other classifiers (i.e., low transferability) is a fundamental problem.  

The transferability of adversarial examples between models implies that when trained with the same dataset, image classifiers (particularly, deep neural networks; DNNs) with different training hyperparameters and architectures end up learning functions that have similar decision boundaries~\cite{goodfellow2014explaining, tramer2017ensemble, moosavi2017universal}. Although the reason behind this training convergence similarity, and thus, the transferability, is not yet fully understood, it is empirically observed that the extent of transferability correlates with the similarity of the network architectures; the more architectural similarity two DNNs share, the more adversarial examples may transfer between them~\cite{tramer2017ensemble, papernot2016transferability}. Therefore, discovering a proper network architecture is also an important factor in order to realize a DNN classifier that attains high clean accuracy as well as low transferability of adversarial examples from other classifiers. Nevertheless, there are only a few studies that investigate and try to alleviate the transferability from the network architecture perspective. Differentiable Architecture Search~(DARTS)~\cite{liu2018darts} has been used to find robust networks but it only works well with weak adversarial attacks~\cite{devaguptapu2020adversarial}. In another study, neuroevolution has been employed to find robust networks through an exhaustive search in a large search space~\cite{kotyan2019evolving}. Several works included adversarial training~\cite{shafahi2019adversarial, wong2020fast} in the network architecture search to improve network robustness and search efficiency. \citet{liu2021multi} incorporated adversarial training to DARTS, which resulted into a more robust network. In addition, some studies have combined oneshot neural architecture search~\cite{cai2019once} with adversarial training to find networks that balances clean and adversarial accuracies~\cite{guo2020meets,xie2021tiny}. Although they are successful in finding networks that reduces transferability, introducing adversarial training can be computationally expensive, especially when using strong adversarial attack methods.

In this paper, we propose a method that discovers network architectures with which one can train a network with low transferability of adversarial examples from a given pretrained network (reference network), while achieving a similar classification accuracy. To find such architectures, we design (i) neuroevolution~\cite{stanley2002evolving}, which yields unconventional network architectures that still perform well~\cite{sun2020automatically}, and introduce (ii) gradient misalignment~(GM), which encourages networks to have dissimilar input gradient behavior from that of the reference network.
The overview of our framework is shown in Fig.~\ref{fig:overview}. 
In the neuroevolution framework, networks evolve over generations; at each generation, networks are architecturally-mutated, trained, and then selected. At the training step, the GM loss is incorporated to reduce the transferability from the reference network. Intuitively, the GM loss measures the dissimilarity of gradient direction at each input points.
As illustrated on Fig.~\ref{fig:overview}, the GM makes it difficult to generate adversarial examples that can deceive both the reference network and evolved network simultaneously; when two networks have largely misaligned input gradients, the direction for adversarially perturbing a clean image to cross the decision boundary of each network is totally different, which prohibits the adversarial example to be effective on both networks.
Note that the idea of exploiting input gradients to obtain different functions (or networks) itself is not new~\cite{jalwana2020orthogonal,kariyappa2019improving, kera2020gradient,kera2021border,kera2021monomial}. However, the existing methods focus on a fixed function architecture, whereas in this study, we are interested in the effect of architecture on obtaining different functions. In the experiments, we show that the GM becomes by far more effective when it is combined with an evolved network architecture rather than using it with the network architecture of the reference network.
It is also worth noting that the neuroevolution designed in this study is more efficient than other neuroevolution-based architecture search in the related studies. \citet{kotyan2019evolving} proposed a powerful evolutionary method which includes block-wise, layer-wise, and model-wise evolution that pursues adversarially robust networks. Although, with considerably large number of generations, their method can discover robust networks, the exhaustive evolution requires high computational cost. In contrast, our neuroevolution framework incorporates GM in evolution, which allows us to efficiently explore architectures with a small number of generations.

In the experiments, we show that the network architecture is actually an important factor to alleviate the transferability of adversarial examples. 
Three standard datasets~(CIFAR-10~\cite{krizhevsky2009learning}, MNIST~\cite{lecun1998gradient}, and KMNIST~\cite{clanuwat2018deep}) are tested in our method.  
Using a pretrained reference network (ResNet-34~\cite{he2016deep}), we generate adversarial examples from the dataset. Then, we perform neuroevolution with GM to evolve networks using the CIFAR-10 dataset. The networks produced are also applied on MNIST and KMNIST datasets to test the versatility of the networks. The results show that for all the datasets, the networks obtained by the proposed framework (neuroevolution with GM) give clean accuracy comparable with that of the reference network while achieving significantly higher adversarial accuracy. Moreover, we demonstrate that training a network with evolved architectures using GM yields a classifier that achieves remarkably better adversarial accuracy than simply training a network with the same architecture as the reference network using GM, which highlights the contribution of the evolved architectures to the adversarial accuracy. The neuroevolution networks with GM maintain the adversarial accuracy even for the adversarial examples generated from other pretrained networks (ResNet-18, VGG~\cite{simonyan2014very}, DenseNet~\cite{huang2017densely}, and SqueezeNet~\cite{iandola2016squeezenet}); whereas other methods significantly drop their adversarial accuracy due to the transferability of adversarial examples across hand-engineered network architectures~\cite{tramer2017ensemble, wu2020skip}. 
To summarize, the experimental results strongly support our hypothesis that the architecture design of networks plays an important role in training networks dissimilar from the reference network and that our framework serves as a powerful tool for discovering such useful network architectures. 

\section{Related work}

\subsection{Transferability of adversarial examples}
Adversarial examples generated to fool a certain network can be transferred to fool other networks. Extensive studies on such transferability, though empirically, revealed numerous findings~\cite{goodfellow2014explaining, biggio2013evasion, dong2018boosting, liu2016delving, moosavi2017universal, papernot2017practical, papernot2016transferability, szegedy2013intriguing, tramer2017space, demontis2019adversarial}. 
First, some studies attribute the transferability to similar decision boundaries, which suggests similarity of functions among the networks~\cite{goodfellow2014explaining, moosavi2017universal, tramer2017space}. 
Second, complex networks has been known to be more susceptible to transferability than simple networks~\cite{demontis2019adversarial}, and the skip connection, which is essential in training very deep networks~\cite{he2016deep}, can be utilized to cause higher transferability~\cite{wu2020skip}.  Third, it has been observed that networks with high transferability between them have aligned gradients~\cite{liu2016delving, demontis2019adversarial} and smoothing the gradients can increase transferability~\cite{wu2020towards}. 
In contrast, some studies reduce transferability by decreasing the magnitude of input gradients~\cite{lyu2015unified, ross2018improving} and by reversing their direction~\cite{jalwana2020orthogonal,kariyappa2019improving}. However, only a limited number of studies, which are published very recently, take network architecture design into account to alleviate transferability~\cite{kotyan2019evolving, devaguptapu2020adversarial,liu2021multi,xie2021tiny,guo2020meets}. Since by observation, the extent of transferability of adversarial examples between two models increases with the architecture similarity,  we believe that architecture search has a good potential in reducing transferability. This study provides a framework to efficiently find unique network architectures that reduce transferability and contribute to the emerging trend in the studies on transferability.

\subsection{Architecture search}
As datasets continue to grow, DNNs have become increasingly complex to learn important features for classification. However, it is not straightforward to hand-engineer an optimal DNN architecture for each task and dataset. Neural Architecture Search~(NAS) algorithms are a line of methods that optimize the network architecture along with the standard parameter optimization~\cite{zoph2016neural, sun2020automatically, cai2019once}. One of the successful NAS algorithms is neuroevolution~\cite{stanley2002evolving} and it has been applied in many different tasks such as image classification, image detection, reinforcement learning etc.~\cite{suganuma2017genetic, operiano2020neuroevolution, sun2020automatically}. With neuroevolution, one can automatically produce unconventional but successful networks that compete with best hand-engineered networks~\cite{sun2020automatically}. In some studies, neuroevolution is used for adversarial defense. In \cite{kotyan2019evolving}, neuroevolution finds networks that are robust against adversarial examples, although an exhaustive search in a large search space is required. Another work has employed a different NAS called Differentiable Architecture Search (DARTS)~\cite{liu2018darts} to find robust networks, which has shown to be effective only to weak adversarial attacks~\cite{devaguptapu2020adversarial}. Several studies have incorporated adversarial training into the architecture search, resulting in less exhaustive search. For example, DARTS yields a more robust network when it is combined with adversarial training~\cite{liu2021multi}. Another example is oneshot NAS~\cite{cai2019once}, which includes adversarial training to find a family of robust networks~\cite{xie2021tiny, guo2020meets}. Although, adversarial training helps improve adversarial example resistance, it still requires significant computational cost, especially when it uses strong adversarial attack method in the training. In this study, we instead propose to exploit input gradients by GM in the architecture search for finding adversarial robust model. GM is more efficient than adversarial training and also requires only a slight modification in the existing training process.

\subsection{Input gradients}
The gradients of the loss of a DNN with respect to the input, or input gradients of a DNN, has been utilized in various ways~\cite{lyu2015unified, ross2018improving, wu2020towards, dong2018boosting, jalwana2020orthogonal}. One of the useful applications of input gradients is to visualize the focused areas on an input image~\cite{simonyan2013deep,sundararajan2017axiomatic,sundararajan2017axiomatic}. It has also been used to generate adversarial examples in standard methods such as Fast Gradient Sign Method (FGSM)~\cite{goodfellow2014explaining} and Projected Gradient Descent (PGD)~\cite{madry2017towards}. Furthermore, the transferability of adversarial examples is enhanced by smoothing the input gradients of a network~\cite{wu2020towards} or by averaging input gradients on an ensemble of networks~\cite{dong2018boosting, liu2016delving, tramer2017space}. Similar direction of input gradients of two networks implies high transferability of adversarial examples between them~\cite{liu2016delving, demontis2019adversarial}. This observation has been recently exploited in the adversarial defense~\cite{kariyappa2019improving,jalwana2020orthogonal}. In~\cite{jalwana2020orthogonal}, given a network, a new network with the same architecture is trained such that its input gradients align differently from those of the given network and thus reduces transferability between them. In this paper, we exploit the input gradients in the same manner as in~\cite{jalwana2020orthogonal} to reduce the transferability of adversarial examples. The critical difference from their study is that for maximizing the gradient misalignment, we dynamically evolve networks through neuroevolution. 

\section{Method}
Here, we describe the process of finding network architectures that reduce transferability of adversarial examples using neuroevolution with GM. Our idea is based on two empirical observations:  (i) the extent of transferability of adversarial examples between networks increases along with their architecture similarity~\cite{tramer2017ensemble, papernot2016transferability}; and (ii) networks with high transferability exhibits aligned input gradients~\cite{liu2016delving, demontis2019adversarial}. 
We first describe GM, to show how input gradients are exploited in the training to reduce transferability and then present overall neuroevolution with GM procedure to evolve networks that converge into functions different from the reference network.

\subsection{Gradient misalignment}\label{meth:GM}
One of the observed characteristics of networks with high transferability is the alignment of their loss gradients with respect to the input, or input gradients~\cite{liu2016delving, demontis2019adversarial}. The input gradients of networks pointing to the same direction imply similar decision boundaries, which makes an adversarial perturbation on an image effective on these networks. Therefore, changing the alignment of the input gradients can change the network decision boundaries where an adversarial example cannot be simultaneously effective on networks with input gradients pointing on opposite directions as shown in Fig.~\ref{fig:overview}. GM encourages the networks to align their input gradients to the opposite direction with respect to the input gradients of a reference network. 

Let $f_c$ and $f_r$ be the network to train and reference network, respectively. Also, let $\ell(\cdot, \cdot)$ be the classification loss (e.g., cross-entropy loss). 
Given a dataset $\mathcal{D}$, the GM loss $\mathcal{L}_{\mathrm{GM}}$, which measures how two input gradients are \textit{misaligned}, is defined as the average cosine similarity of gradients of $f_c$ and $f_r$ as follows.
\begin{align}
\label{eq:gm_loss}
    \mathcal{L}_{\mathrm{GM}} &= \frac{1}{|\mathcal{D}|}\sum_{(x,y)\in\mathcal{D}} \frac{\langle\nabla_x\ell(f_c(x),y),\nabla_x\ell(f_r(x),y)\rangle}{\|\nabla_x\ell(f_c(x),y)\|,\|\nabla_x\ell(f_r(x),y)\| },
\end{align}
where $|\mathcal{D}|$ denotes the size of dataset, $\langle\cdot, \cdot\rangle$ denotes the inner product of vectors, and $\|\cdot\|$ is the Euclidean norm.

Combining the GM loss with a standard cross-entropy loss, we can train a network to achieve high clean accuracy and low transferability from the reference network. Thus, the overall loss becomes
\begin{align}
\label{eq:total_loss}
    \mathcal{L} = \mathcal{L}_{\mathrm{CE}} + \lambda \mathcal{L}_{\mathrm{GM}},
\end{align}
where $\lambda$ is a hyperparameter. The hyperparameter $\lambda$ is carefully adjusted depending on the characteristics of a dataset. On one hand, if the $\lambda$ is too big, the network fails to learn the correct labels of images because it simply focuses on input gradients misalignment. On the other hand, if the $\lambda$ is too small, the network fails to encourage input gradients misalignment.

\subsection{Neuroevolution}
The neuroevolution framework designed in this work borrows some ideas from steady-state genetic algorithm~\cite{vargas2016spectrum, agapie2014theoretical}. Unlike typical neuroevolution, where at each generation, the population is created from a batch of selected and mutated candidate networks, our framework at each generation replaces a candidate network only when it is outperformed by one of the children networks. With this idea, we can ensure that each candidate network always improve along the generations. We can also save computational cost because it does not require a large population size to succeed unlike in the typical method. In particular, we only keep a few candidate networks (e.g., four or five networks) in a population and each network produces two to three children only. In our experiments, we empirically observed that such a small number of networks and children are sufficient to produce networks that substantially reduce transferability.

Before the start of neuroevolution with GM method, a pretrained reference network is prepared, which is the baseline for clean accuracy and adversarial accuracy improvement. It is also used to produce the adversarial examples equivalent of all the test set images for the adversarial accuracy evaluation. 

The basic neuroevolution framework is composed of network mutation, training, fitness evaluation, and selection, which repeats through generations. Specifically, in our neuroevolution with GM method, at each generation, the following procedures are performed on each candidate network in the population: 
\begin{enumerate}[label=Step \arabic*., leftmargin=*]
    \item A candidate network in the population produces children by mutation. The children are trained with GM with respect to the reference network.
    \item Each child is evaluated for fitness and compared to the architecturally closest candidate network in the population.
    \item The children with better fitnesses replace their closest candidate network counterparts.  
\end{enumerate}

At initialization, the candidate networks are generated to have diverse architectures. Because there are limited number of candidate networks in a population, it is important to make their architectures diverse as much as possible. Having a diverse set of candidate networks lets neuroevolution explore more areas in the search space and find more potential solutions. To this end, we adopt a modified Spectrum-based Niching~\cite{vargas2016spectrum}. Each network is represented by a vector, called \textit{spectrum}, which contains the following convolutional neural network properties: number of convolutional blocks, number of pooling blocks, number of strided convolutional blocks, number of summation blocks, and number of concatenation blocks. The distance between two network architectures is measured using the Euclidean distance between their spectrums. Formally, the distance of networks $N_1, N_2$ is defined by
\begin{align}
d_{\mathrm{spec}}(N_1,N_2) &= \| \mathrm{spec}(N_1) - \mathrm{spec}(N_2)\|, 
\end{align}
where $\mathrm{spec}(N_i)$ denotes the spectrum of network $N_i$.

In Step~1, each candidate network is mutated by adding, editing, or deleting a convolutional block, a pooling block, or a strided convolutional block. In addition, summation block, which sums the output channels of two prior blocks, and concatenation block, which concatenates the output channels of two prior blocks, are added as skip connection blocks. In the summation block, if the two input blocks do not have the same output channels, the smaller output channel is padded with zeros to match the larger output channel before summing. To facilitate effective architecture search, aggressive mutation is employed by applying multiple mutations to a candidate network to produce a child.

In Step~2, the children networks are trained with GM as explained in the previous Section~\ref{meth:GM} and they are evaluated using a fitness function. The fitness is based on both accuracy on clean images (i.e., clean accuracy) and accuracy on the transferred adversarial examples (i.e., adversarial accuracy) in order not to reduce the transferability at the cost of clean accuracy. Specifically, the fitness of a network is defined by the minimum of the two accuracies.  
\begin{align}
\label{eq:fitness}
   \mathcal{F} &= \min\{\mathcal{A}_{\mathrm{CL}}, \mathcal{A}_{\mathrm{AD}}\},
\end{align}
where $\mathcal{A}_{\mathrm{CL}}$ is the clean accuracy and $\mathcal{A}_{\mathrm{AD}}$ is the adversarial accuracy.
By using the minimum between the clean accuracy and adversarial accuracy as the fitness, we ensure that the lower bound of clean accuracy and adversarial accuracy will always increase along the course of evolution. 

\begin{figure*}[t]
\centerline{\includegraphics[scale=0.22]{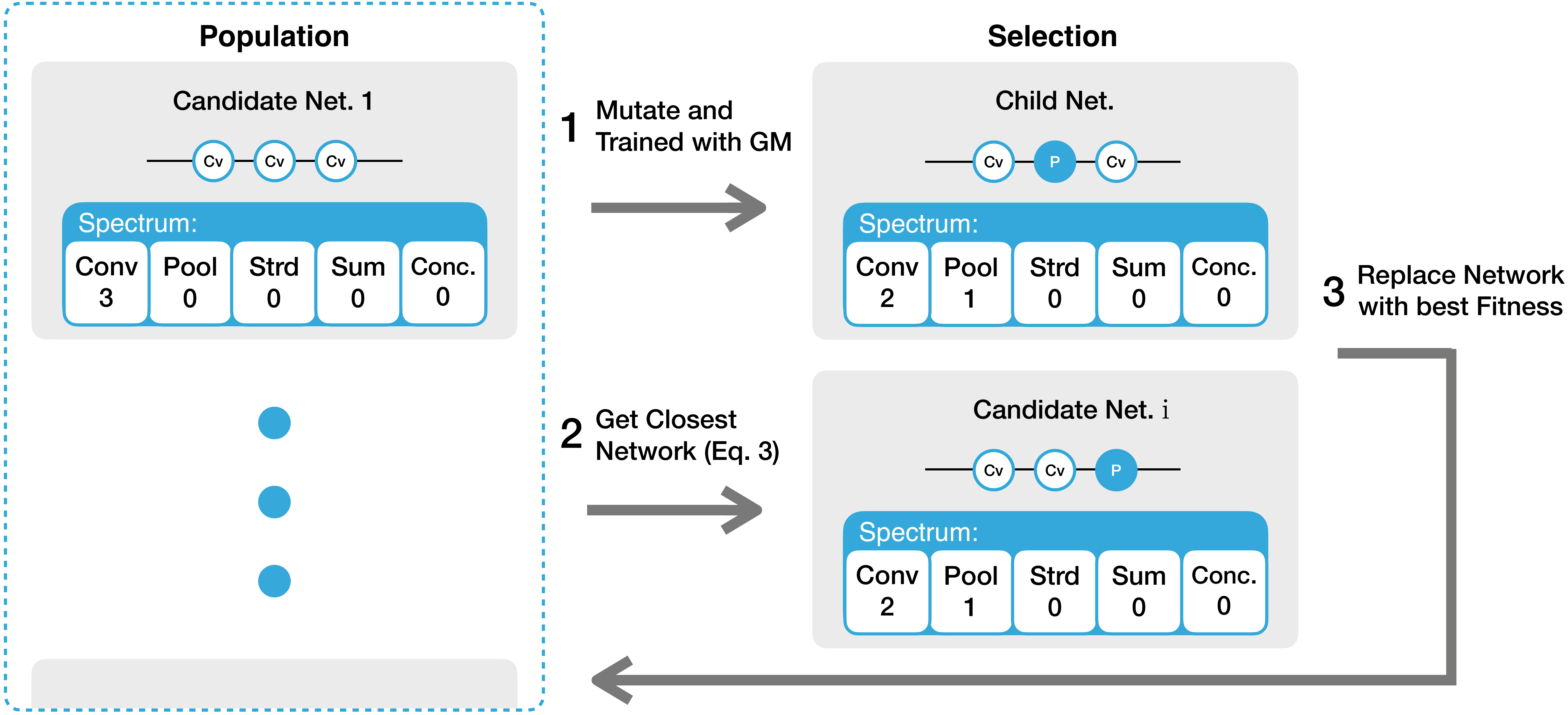}}
\caption{The \textbf{selection} process is demonstrated with candidate network 1 where it produces a child. The child is trained with GM and compared to the closest candidate network in terms of architecture using the network distance (i.e., Euclidean distance of spectrums). If the fitness of the child is better than the closest candidate network, it replaces that candidate network in the population.}
\label{fig:mutation and selection}
\end{figure*}

In Step~3, the spectrum of each child from every candidate network is obtained. Afterwards, the closest candidate network to a specific child is taken by comparing the child spectrum to every candidate network spectrum and getting the smallest network distance. The fitnesses of the child and candidate network pair are compared. If the child has a better fitness, it replaces the candidate network in the population. This selection process using the spectrum is illustrated in Fig.~\ref{fig:mutation and selection}

\begin{algorithm}[t]
 \caption{Neuroevolution with GM}
 \begin{algorithmic}[1]
 \renewcommand{\algorithmicrequire}{\textbf{Input:}}
 \renewcommand{\algorithmicensure}{\textbf{Output:}}
 \REQUIRE $s$: size of population, $n$: number of children to produce, $d$: minimum network distance, $g$: number of generations  
 \ENSURE $F$: evolved networks after neuroevolution \\
  \STATE $P = \{N_1, N_2, ..., N_{s}\}$ \quad \COMMENT \ Initialize the population by $s$ candidate networks $N_1, \ldots, N_{s}$ with minimum network distance $d$.
  \FOR {$t = 1, 2, \ldots, g$}
    \FOR {\textbf{each} $N$ \textbf{in} $P$}
        \STATE $\{C_1, C_2, \ldots, C_{n}\}= \mathrm{mutation}(N)$
        \FOR{$j = 1, 2, \ldots, n$}
                \STATE $N_j= \mathrm{train}(C_j)$ \quad \COMMENT \ Train with loss Eq.~(\ref{eq:total_loss}). 
            \STATE $N^{*} = \argmin_{N^{\prime} \in P} d_{\mathrm{spec}}(N^{\prime}, C_j)$ \quad \COMMENT \ Find the closest network
            \IF {$\mathrm{\mathcal{F}}(N^{*}) < \mathcal{F}(C_j)$} \STATE Replace $N^{*}$ with $C_j$.
            \ENDIF
        \ENDFOR
    \ENDFOR
  \ENDFOR
 \RETURN $P$
 \end{algorithmic}
 \end{algorithm}

The neuroevolution with GM method is summarized in Algorithm 1. Given the population size $s$, number of children $n$ (of each candidate network), network distance value $d$, and number of generations $g$ as inputs, a population $P$ is initialized with candidate networks with a minimum network distance of $d$ to each other. For every generation in $g$, each candidate network in $P$ produces $n$ children through aggressive mutation. Afterwards, each child in $n$ is trained with GM and its fitness is compared to the candidate network in current $P$ with the smallest network distance. If there are multiple candidate networks with the smallest network distance, the first closest network is used. Although, there is a low probability that this will occur due to the combination of aggressive mutation and spectrum-based niching. During the comparison, if the child has better fitness, it replaces that particular candidate network in the current $P$. At the end of the neuroevolution process, the population of evolved networks $P$ is returned as output $F$. The evolved networks $F$ are trained further for refinement.

\section{Experiments}

\noindent\textbf{NE+GM networks and baselines.} 
In the experiments, we demonstrate that, exploiting neuroevolution~(NE) and GM, the proposed framework can produce architectures that attain both high clean accuracy and high adversarial robustness against adversarial examples transferred from the reference network. We compare four networks: (i) two NE+GM networks, which refers to the top two networks produced by our framework; (ii) reference network, which is the network used to generate adversarial examples; and (iii) reference network with GM, which is a network with the architecture of the reference network and trained with GM. The networks (ii) and (iii) are denoted as baseline networks. In Section~\ref{subsec:Reduced-Dataset}, we introduce a hand-engineered network as another baseline.

\noindent\textbf{Implementation details.} In the experiments, the NE population is initialized with four or five candidate networks that are randomly designed to have at least a network distance of four to each other. The batch size is fixed to 128 and the other hyperparameters (e.g., learning rate) are kept in default PyTorch DNN library~\cite{paszke2019pytorch} settings without any fine-tuning to establish objective comparison between the baseline networks and NE+GM networks. To obtain NE+GM networks, NE runs for 50 generations and in every generation, each candidate network produces two children networks using aggressive mutation (i.e., mutation for four or five times to produce a child). The children are trained with GM for 20 epochs and the number of epochs increases by five every 10 generations to account for the network complexity that grows every generation. This implementation also helps decrease training cost and time considering that simple networks can converge at small epochs. After completing the whole process of NE, the candidate networks in the final population are given as NE+GM networks after they are refined by additional training for another 1,000 epochs. This ensures that the clean and adversarial accuracies of NE+GM networks converge to a stable value.

\noindent\textbf{Outline of experiments.} We demonstrate the effectiveness of the proposed framework by two experiments: (i) full-dataset experiment and (ii) reduced-dataset experiment. In experiment (i), we use full CIFAR-10~\cite{krizhevsky2009learning}, MNIST~\cite{lecun1998gradient}, and KMNIST~\cite{clanuwat2018deep} as datasets and ResNet-18~\cite{he2016deep}, VGG~\cite{simonyan2014very}, DenseNet~\cite{huang2017densely}, and SqueezeNet~\cite{iandola2016squeezenet} as network models. 
We evaluate NE+GM networks and baselines using the datasets as well as adversarial examples generated by the reference network and the aforementioned four network models. We demonstrate that the NE+GM networks notably outperforms the baseline networks in terms of the adversarial accuracy while maintaining a good clean accuracy. 
We are also interested in whether the proposed framework remains effective for small datasets because there exists various applications where only limited data are available~\cite{operiano2020neuroevolution,shaikhina2017handling,zhang2018strategy,oh2020deep,wang2020efficient}.  
To this end, in experiment (ii), we use the significantly reduced versions of the aforementioned datasets and architecturally small hand-engineered networks as baseline networks to compensate for the limited dataset. We show that even when the dataset size is limited, the proposed framework can find good architectures that can attain high clean accuracy and good adversarial accuracy as in the full-dataset experiment. In experiment (ii), we also perform additional analysis to confirm two observations. First, the architectures of NE+GM networks of some limited dataset are still useful even when they are trained for other datasets. Second, NE+GM networks outperforms the reference network even when it is combined with several standard defense methods.

\subsection{Full-dataset experiment}

\subsubsection{Clean accuracy and adversarial accuracy}\label{subsec:accuracies-full}
\begin{table*}[t]\centering
\footnotesize
\ra{1.3}
\caption{The clean accuracy and adversarial accuracy of the baseline networks and NE+GM networks trained with a full dataset is shown here. In this table, full CIFAR-10 dataset is employed. The NE+GM networks have remarkably higher adversarial accuracy (i.e., lower transferability of adversarial examples) than the reference network without GM (Ref. Network) and with GM (Ref. Network+GM).}
\begin{tabular}{@{}lcccccc@{}}\toprule

& Ref. Net. & Ref. Net.+GM & NE+GM Net. 1 & NE+GM Net. 2 \\ \midrule
Clean Acc. & 78.40\% & 80.82\% & \textbf{81.09\% } & 80.62\% \\
Adv. Acc. ($L_\infty$-PGD) & 23.10\% & 42.90\% & \textbf{54.12\%} & 53.27\%  \\
\bottomrule
\label{table:accuracies full dataset}
\end{tabular}
\end{table*}

We focus on the results with the full CIFAR-10 dataset. Similar results are obtained for other datasets~(cf. Section~\ref{subsec:other-datasets}).
We use a ResNet-34 as the reference network. Using the reference network, we generate adversarial examples for each test image in CIFAR10 using PGD with $L_{\infty}$ norm ($L_{\infty}$-PGD.  Table~\ref{table:accuracies full dataset} reports the results on clean accuracy and adversarial accuracy of networks of the proposed methods (NE+GM networks) and baselines. All of the NE+GM networks have slightly higher clean accuracy than the reference network. 
Noticeably, the NE+GM networks shows remarkable increase in the adversarial accuracy; they outperform the reference network by 31\%. It is worth noting that the NE+GM networks also outperform the reference network with GM by at least 11\%, while maintaining a good clean accuracy. This indicates that NE+GM successfully discovers robust network architectures which make it easier to misalign input gradients by GM, resulting in lower transferability of adversarial examples from the reference networks. 

\begin{figure}[t]
\centerline{\includegraphics[scale=.28]{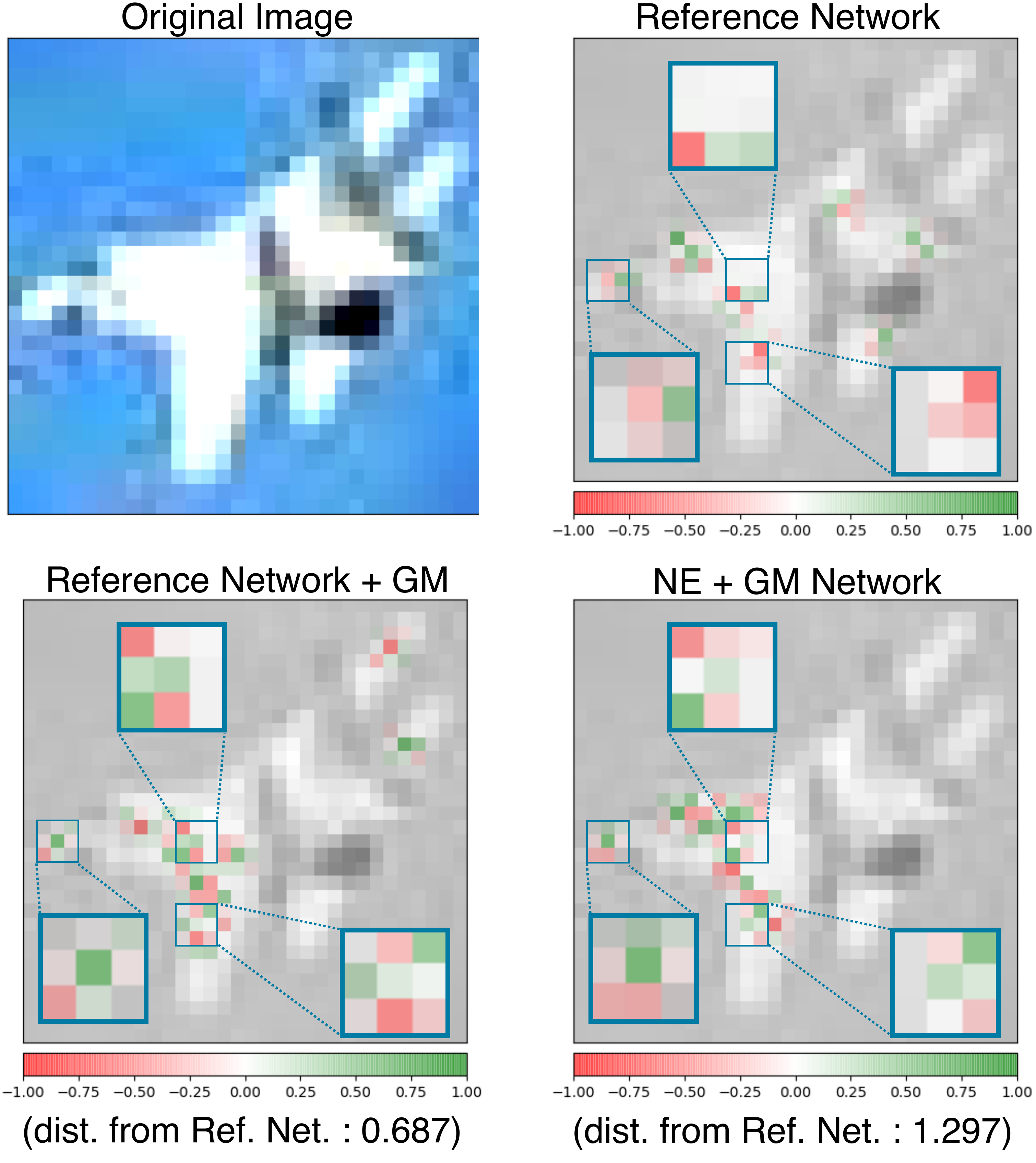}}
\caption{The integrated gradients~\cite{sundararajan2017axiomatic} of the reference network, reference network with GM, and NE+GM network show the focus areas of each network using the relative values ranging in $[-1,1]$. 
By contrasting colors of the same pixel locations, one can see that the focus area of reference network with GM (lower left) and NE+GM network (lower right) are both dissimilar to that of the reference network. The total pixel distance of the NE+GM network (1.297) are higher than that of the reference network with GM (0.687). This indicates that the input gradient of the NE+GM network is more dissimilar from that of the reference network.
}
\label{fig:integratedgradients}
\end{figure}

The effects of GM on networks can be visually explained using integrated gradients~\cite{sundararajan2017axiomatic}. In Fig.~\ref{fig:integratedgradients}, we provide the input gradient value on each pixel of three networks namely: reference network, reference network with GM, and NE+GM network. Red pixels and green pixels mean negative and positive direction respectively. One can see inside the upper right blue box that the color of the pixels in reference network becomes the opposite in the reference network with GM and NE+GM network. The change in color indicates the reversal in the direction of input gradients. The same shift in input gradients direction can also be said on other boxes. This implies that GM can certainly encourage networks to have misaligned input gradients with respect to the reference network. Furthermore, when the total pixel distance with the reference network is calculated, the NE+GM network has a higher value than the reference network with GM with 1.297 to 0.687 respectively. Hence, the proposed framework can find better architectures that can misalign input gradients more than the architecture of the reference network.

\subsubsection{Transferability of adversarial examples generated from other networks}
\begin{table*}[t]\centering
\footnotesize
\ra{1.3}
\caption{The adversarial robustness of the baseline networks and NE+GM network against adversarial examples from different networks are displayed here. NE+GM network shows consistent robustness while two baselines do not. } \vspace{1mm}
\begin{tabular}{@{}lcccccc@{}}\toprule

& Ref. Network & Ref. Network+GM & NE+GM Network \\ \midrule
ResNet-18 & 36.84\% & 47.40\% & \textbf{53.99\% }\\
VGG & 42.98\% & 52.35\% & \textbf{55.23\%}  \\
DenseNet & 36.67\% & 48.44\% & \textbf{53.93\%}  \\
SqueezeNet & 44.23\% & 52.48\% & \textbf{55.09\%}  \\
\bottomrule
\label{table:transferability networks}
\end{tabular}
\end{table*}

The NE+GM networks are also robust against adversarial examples generated using other networks in addition to the reference network. 
Here, four sets of adversarial examples are generated from CIFAR-10 test set using four other standard networks (ResNet-18, VGG, DenseNet, and SqueezeNet) instead of the reference network~(ResNet-34). 
As shown in Table~\ref{table:transferability networks}, for any set of adversarial examples, the reference network is the least adversarially robust (i.e., high transferability). The reference network with GM exhibits better robustness by 5-10\% than the reference network, which indicates that GM reduces transferability. Due to the architectural superiority, the NE+GM network achieved the best adversarial accuracies (the lowest transferability). 
One may notice that the extent of transferability varies depending on the type of network architectures. For example, the reference network shows lower transferability of adversarial examples from VGG than adversarial examples from the ResNet-18 and DenseNet. This is because the VGG network does not have skip connections while the reference network~(ResNet-34) utilizes skip connections. The reference network has also lower transferability from SqueezeNet because SqueezeNet does not employ skip connections as ResNet and DenseNet do.
In contrast, the NE+GM network, even with skip connections, is specifically designed to have a distinct architecture that reduces transferability from the reference network. 
Therefore, it can reduce transferability from networks similar to the reference network and also networks that do not have skip connections. 

\subsubsection{Results in other datasets}\label{subsec:other-datasets}
\begin{table*}[t]\centering
\footnotesize
\ra{1.3}
\caption{The baseline networks and the NE+GM network produced in Section~\ref{subsec:accuracies-full} are retrained using MNIST and KMNIST datasets. Among the networks, the NE+GM network has the best clean accuracy and adversarial accuracy.}\vspace{1mm}
\begin{tabular}{@{}lccccc@{}}\toprule
& \multicolumn{3}{c}{MNIST} \\
\cmidrule{2-4}  
& Ref. Net. & Ref. Net.+GM & NE+GM Net.  \\ \midrule
Clean Acc. & 97.71\% & 97.73\% & \textbf{99.10\%} \\
Adv. Acc.($L_\infty$-PGD) & 87.90\% & 95.50\% & \textbf{98.10\%} \\
\midrule & \multicolumn{3}{c}{KMNIST} \\
\cmidrule{2-4}  
& Ref. Net. & Ref. Net.+GM & NE+GM Net.  \\ \midrule
Clean Acc & 95.69\% & 96.06\% & \textbf{97.51\%} \\
Adv. Acc.($L_\infty$-PGD) & 75.80\% & 89.10\% & \textbf{94.40\%} \\
\bottomrule
\label{table:different full datasets}
\end{tabular}
\end{table*}

Here, the NE+GM network produced in Section~\ref{subsec:accuracies-full} is retrained using full MNIST and KMNIST datasets (i.e., trained from scratch) independently, and still achieved notable clean accuracy and adversarial accuracy for each dataset.
As seen in Table~\ref{table:different full datasets}, the results show better clean and adversarial accuracies from the NE+GM network, which agree with the results in Table~\ref{table:accuracies full dataset}. Despite not evolving a new NE+GM network specialized to these datasets, the NE+GM network has still achieved higher accuracies than the reference network with GM on all comparisons. This performance is attributed to the NE+GM architecture created for the full CIFAR-10 dataset, which provided abundant features to learn from. Consequently, the well-developed NE+GM network architecture can adjust to the features of different datasets (i.e., MNIST and KMNIST) and perform well even if it is not designed for those.

\subsection{Reduced-Dataset experiment}\label{subsec:Reduced-Dataset}

\subsubsection{Clean accuracy and adversarial accuracy}
\begin{table*}[t]\centering
\footnotesize
\ra{1.3}
\caption{The clean accuracy and adversarial accuracy of baseline networks, an additional baseline network hand-engineered network with GM (denoted as Hand Eng. Network+GM), and NE+GM network trained with a reduced dataset (i.e., reduced CIFAR-10) is shown here. The results show that NE+GM networks has the best clean accuracy while having comparable adversarial accuracy with reference network with GM.}
\begin{tabular}{@{}lcccccc@{}}\toprule
& Ref. Net. & Ref. Net.+GM & HE. Net.+GM & NE+GM Net. 1 & NE+GM Net. 2 \\ \midrule
Clean Acc. & 50.00\% & 53.40\% & 55.40\% & \textbf{57.80\% }& 56.20\% \\
Adv. Acc. & 8.80\% & 63.06\% & 43.32\% & 63.17\% & \textbf{63.56\%}  \\
\bottomrule
\label{table:accuracies reduced dataset}
\end{tabular}
\end{table*}

In this section, we mainly discuss the results on the reduced CIFAR-10 dataset. The results in other dataset is reported in Section~\ref{subsec:other-red-datasets}. To obtain the reduced CIFAR-10 dataset, the full CIFAR-10 training set and test set are cut down to 10,000 training images (1,000 images per label) and 1,000 test images (100 images per label) respectively. Accordingly, we generate the adversarial examples for each image in the reduced test set using the reference network and $L_\infty$-PGD. 
Here, the reference network is a simple hand-engineered network because a limited dataset cannot effectively train a deep network such as ResNet-34.
In addition to the baseline networks (reference network trained with and without GM), another different simple hand-engineered network with GM is introduced as additional network to compare. The simple hand-engineered networks are each composed of convolutional blocks, pooling blocks, and skip connections that are not more than ten blocks in total (see Appendix \ref{apd:simple}).

As shown in Table~\ref{table:accuracies reduced dataset}, all of the networks trained with GM have achieved better clean accuracy than the reference network. In particular, the NE+GM networks have better clean accuracy than all the other networks, which confirm the ability of the proposed framework to discover better network architectures. It is also worth noting that while NE+GM networks and reference network with GM have comparable adversarial accuracies, the adversarial accuracies are higher than their clean accuracies. This result is counter-intuitive but the same behavior is not observed in the full dataset experiment. Thus, we consider this behavior to be specific to reduced datasets. Finally, the higher adversarial accuracy of NE+GM networks compared to the hand-engineered network+GM shows that the extent of transferability can be reduced by a properly designed network architecture.

\subsubsection{The impact of the choice of adversarial attack methods}\label{subsec:different-attacks}
\begin{table*}[t]\centering
\footnotesize
\ra{1.3}
\caption{The fooling rate under three adversarial attack methods. (Lower is better.) NE+GM networks perform best for most attack methods.}
\begin{tabular}{@{}lccccc@{}}\toprule

& Ref. Net.+GM & HE. Net.+GM & NE+GM Net. 1 & NE+GM Net. 2 \\ \midrule
$L_\infty$-PGD & 17.42\% & 30.26\% & \textbf{16.10\%} & 18.77\% \\
$L_2$-PGD & 36.04\% & 36.41\% & \textbf{28.83\%} & 31.19\%  \\
FGSM & 69.07\% & \textbf{54.36\%} & 60.26\% & 65.21\%  \\
\bottomrule
\label{table:fool ratio attack}
\end{tabular}
\end{table*}

Here, we see the impact of the choice of attack method on the performance of the baseline networks and NE+GM networks produced in Section~\ref{subsec:Reduced-Dataset}.
We use PGD with $L_\infty$ norm, PGD with $L_2$ norm ($L_2$-PGD), and FGSM to generate adversarial examples on the reduced CIFAR-10 test set. 
We employ fooling rate as a convenient evaluation metric because with the limited data, the clean accuracy tends to be relatively low and fooling rate can quantify the behavior of networks to adversarial examples more clearly based on the ratio of images that change labels when adversarially attacked. For each adversarial attack method, the fooling rate of every network in the baseline and NE+GM networks is calculated by extracting all the clean images that are correctly classified by both the reference network and the network that is being tested. Afterwards, the extracted images are adversarially attacked using the reference network and the adversarial attack method to generate adversarial examples. Then, the adversarial examples are evaluated by the network that is being tested. 
The results reported in Table~\ref{table:fool ratio attack} show that the fooling rate of networks on $L_\infty$-PGD agrees with the previous adversarial accuracy results on Table~\ref{table:accuracies reduced dataset} as expected. Moreover, since $L_2$-PGD is known to produce weaker adversarial examples compared to $L_\infty$-PGD, using GM on the networks results into a lower fooling rate on $L_\infty$-PGD than $L_2$-PGD due to more vulnerable input gradients of the reference network to $L_\infty$-PGD. The stronger the adversarial attack is, the more effective GM becomes in reducing transferability. However, among the networks, the NE+GM network still has the lowest fooling rate for $L_2$-PGD. FGSM is the weakest adversarial attack among the three adversarial attacks. Thus, the FGSM perturbation level or $\epsilon$ is adjusted to create stronger attacks despite with more obvious image perturbations for easier comparison between networks. Unexpectedly, the hand-engineered network+GM has the lowest fooling rate. One of the reasons it has a lower fooling rate than the NE+GM network, is because the adversarial examples used in the NE evolution are generated using $L_\infty$-PGD. Although, the NE+GM networks are still better than reference network with GM.

\subsubsection{Comparison with standard adversarial defense}
\begin{table*}[t]\centering
\footnotesize
\ra{1.3}
\caption{The fooling rate of the baseline networks and NE+GM networks are compared to different defense methods. The free adversarial training and fast adversarial training are denoted as A.T. (Free) and A.T. (Fast) respectively in this table. The results show that NE+GM networks can perform better than other different defense methods. (Lower is better.)}
\begin{tabular}{@{}lcccccc@{}}\toprule

& Ref. Net. & Ref. Net.+GM & HE. Net.+GM & NE+GM Net. \\ \midrule
$L_\infty$-PGD & 85.00\% & 22.80\% & 24.20\% & \textbf{22.40\%}  \\ \midrule
& JPEG Comp.~\cite{guo2017countering} & Bilateral Filt.~\cite{xie2019feature} & A.T. (Free)~\cite{shafahi2019adversarial} & A.T. (Fast)~\cite{wong2020fast} \\ \midrule
$L_\infty$-PGD & 84.4\% & 48.80\% & 51.00\% & 43.80\% \\
\bottomrule
\label{table:fool ratio defense}
\end{tabular}
\end{table*}

Here, we show that the NE+GM networks can perform better than the reference network combined with several adversarial defense methods. 
We employ two image filtering methods (i.e., JPEG compression~\cite{guo2017countering} and bilateral filtering~\cite{xie2019feature}) and two adversarial training methods (i.e., free adversarial training~\cite{shafahi2019adversarial} and fast adversarial training~\cite{wong2020fast}) as the defense methods. 
Again, fooling rate is used for evaluation. 
In contrast to Section~\ref{subsec:different-attacks}, the adversarial examples here are obtained by extracting the correctly classified clean images using the reference network only and generating their corresponding adversarial examples using $L_\infty$-PGD. The same adversarial examples are used for all the networks tested. Note that the fooling rate defined here is modified to compensate for image filtering methods, which cannot classify images. For the image filtering and adversarial training methods, we use the reference network. 
The results in Table~\ref{table:fool ratio defense} show that image filtering techniques do not offer strong adversarial defense as the NE+GM network due to its passive procedure. The advantage of our proposed NE+GM method is that it actively searches for architectural solutions, which makes it effective even with a strong adversarial attack such as $L_\infty$-PGD. Fast and Free adversarial training methods are good adversarial defense techniques. However, due to the limited dataset employed, balancing the clean accuracy and adversarial accuracy using adversarial training has produced fooling rates that are notably higher than the reference network with GM and NE+GM networks.

\subsubsection{Results in other datasets}\label{subsec:other-red-datasets}
\begin{table*}[t]\centering
\footnotesize
\ra{1.3}
\caption{The baseline networks and the NE+GM network produced in Section~\ref{subsec:Reduced-Dataset} are retrained using other datasets (i.e., MNIST, FMNST, and KMNIST dataset). Although the results between the reference network with GM and NE+GM network are comparable in FMNIST and KMNIST dataset, NE+GM networks consistently performs well on clean accuracy and adversarial accuracy across all the datasets.}\vspace{1mm}
\begin{tabular}{@{}lccccc@{}}\toprule
& \multicolumn{3}{c}{MNIST} \\
\cmidrule{2-4}  
& Ref. Net. & Ref. Net.+GM & NE+GM Net.  \\ \midrule
Clean Acc. & \textbf{93.20\%} & 92.30\% & 89.10\% \\
Adv. Acc.($L_\infty$-PGD) & 63.40\% & 77.00\% & \textbf{84.50\%} \\
\midrule & \multicolumn{3}{c}{FMNIST} \\
\cmidrule{2-4}  
& Ref. Net. & Ref. Net.+GM & NE+GM Net.  \\ \midrule
Clean Acc & \textbf{88.90\%} & 85.90\% & 87.00\% \\
Adv. Acc.($L_\infty$-PGD) & 51.20\% & \textbf{82.90\%} & 81.90\% \\
\midrule & \multicolumn{3}{c}{KMNIST} \\
\cmidrule{2-4}  
& Ref. Net. & Ref. Net.+GM & NE+GM Net.\\ \midrule
Clean Acc. & 85.40\% & \textbf{86.80\%} & 86.60\%\\ 
Adv. Acc.($L_\infty$-PGD) & 56.90\% & 83.50\% & \textbf{83.70\%}\\  
\bottomrule
\label{table:different reduced dataset}
\end{tabular}
\end{table*}

In this experiment, we show that by retraining the NE+GM network produced in Section~\ref{subsec:Reduced-Dataset} with other reduced datasets,  we can attain good clean accuracy and adversarial accuracy with the retrained networks. We use datasets, MNIST, FMNIST~\cite{xiao2017fashion}, and KMNIST and reduce them with the same training and test set count as the reduced CIFAR-10. 
As seen in Table~\ref{table:different reduced dataset}, the reference network has the best accuracies in MNIST and FMNIST in exchange of worst adversarial accuracy. However, both the reference network with GM and NE+GM network manage to balance the clean accuracy and adversarial accuracy except for the MNIST result in which the NE+GM network has a notably higher adversarial accuracy with slightly lower clean accuracy than the reference network with GM. The performance of NE+GM network is attributed to its architecture designed for the reduced CIFAR-10, which has limited size and features. Unlike the NE+GM network in Table~\ref{table:different full datasets}, where the architecture is well-developed, the NE+GM network here is relatively unrefined. 

\section{Conclusion}
In this paper, we addressed from architectural perspective a problem of discovering DNN classifiers that have low transferablity against adversarial examples of other models. 
We proposed a method to evolve DNN architectures to find such networks. In particular, we employed NE to produce unconventional network architectures and GM to encourage input gradients misaligned across networks so that dissimilar functions are learned, which is expected to lead to low transferablity. In the experiments, the proposed method successfully discovered networks with expected properties; they achieved comparable clean accuracy with the reference network as well as significantly lower transferable against the adversarial examples generated from the reference network. Importantly, the networks discovered by the proposed method outperformed the reference network even when it was trained with GM. The discovered networks were also robust to adversarial examples of other models~(ResNet-18, VGG, DenseNet, and SqueezeNet), whereas the reference network~(ResNet-34), trained with or without GM, was vulnerable to them. These results indicate that evolving network architectures played an important role in reducing the transferability.  We consider that our study contributes to a fundamental problem of training adversarially robust classifiers, particularly from the transferability and novel network architecture perspective, which is a promising avenue to explore. 

\appendix
\section{Simple hand-engineered networks}\label{apd:simple}
\begin{figure}[t]
\centerline{\includegraphics[scale=0.25]{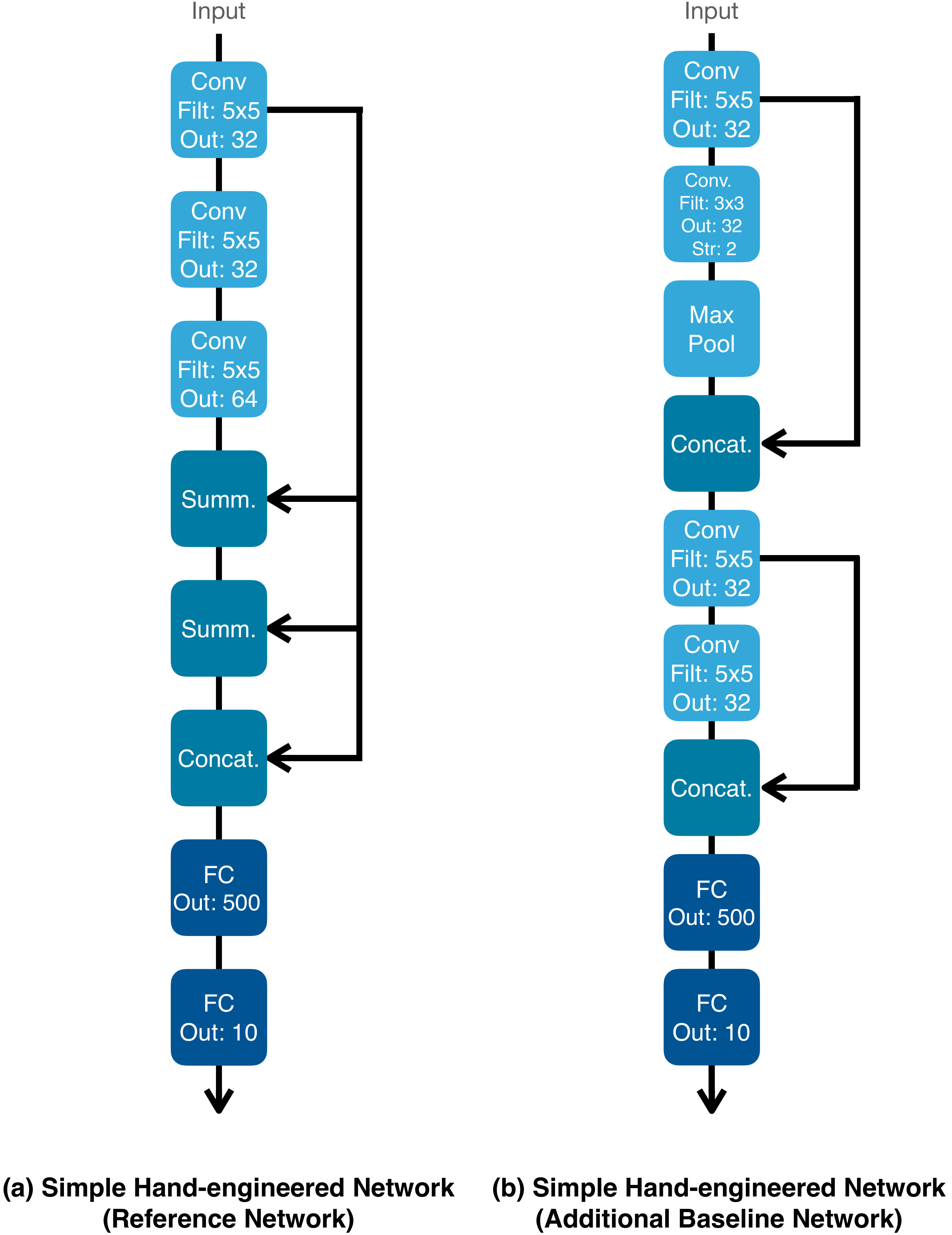}}
\caption{These are the architecture details of the simple hand-engineered networks for the reduced dataset. The network in (a) shows the simple hand-engineered network used as a reference network and the network in (b) shows the simple hand-engineered network used as additional baseline network.}
\label{fig:simple-hand}
\end{figure}

In Fig.~\ref{fig:simple-hand}, we describe the architecture details of two simple hand-engineered networks in the reduced datatset. The network in (a) is employed as the reference network. It has three convolutional blocks, three skip connection blocks (i.e., summation and concatenation blocks), and two fully-connected blocks. Meanwhile, the network in (b) is utilized as an additional baseline network. It has four convolutional blocks, one pooling block, two skip connection blocks, and two fully-connected blocks. 


\begin{thebibliography}{54}
\providecommand{\natexlab}[1]{#1}
\providecommand{\url}[1]{#1}
\csname url@samestyle\endcsname
\providecommand{\newblock}{\relax}
\providecommand{\bibinfo}[2]{#2}
\providecommand{\BIBentrySTDinterwordspacing}{\spaceskip=0pt\relax}
\providecommand{\BIBentryALTinterwordstretchfactor}{4}
\providecommand{\BIBentryALTinterwordspacing}{\spaceskip=\fontdimen2\font plus
\BIBentryALTinterwordstretchfactor\fontdimen3\font minus
  \fontdimen4\font\relax}
\providecommand{\BIBforeignlanguage}[2]{{%
\expandafter\ifx\csname l@#1\endcsname\relax
\typeout{** WARNING: IEEEtranN.bst: No hyphenation pattern has been}%
\typeout{** loaded for the language `#1'. Using the pattern for}%
\typeout{** the default language instead.}%
\else
\language=\csname l@#1\endcsname
\fi
#2}}
\providecommand{\BIBdecl}{\relax}
\BIBdecl

\bibitem[Goodfellow et~al.(2015)Goodfellow, Shlens, and
  Szegedy]{goodfellow2014explaining}
I.~Goodfellow, J.~Shlens, and C.~Szegedy, ``Explaining and harnessing
  adversarial examples,'' in \emph{International conference on learning
  representations}, 2015.

\bibitem[Biggio et~al.(2013)Biggio, Corona, Maiorca, Nelson, {\v{S}}rndi{\'c},
  Laskov, Giacinto, and Roli]{biggio2013evasion}
B.~Biggio, I.~Corona, D.~Maiorca, B.~Nelson, N.~{\v{S}}rndi{\'c}, P.~Laskov,
  G.~Giacinto, and F.~Roli, ``Evasion attacks against machine learning at test
  time,'' in \emph{Joint european conference on machine learning and knowledge
  discovery in databases}.\hskip 1em plus 0.5em minus 0.4em\relax Springer,
  2013, pp. 387--402.

\bibitem[Dong et~al.(2018)Dong, Liao, Pang, Su, Zhu, Hu, and
  Li]{dong2018boosting}
Y.~Dong, F.~Liao, T.~Pang, H.~Su, J.~Zhu, X.~Hu, and J.~Li, ``Boosting
  adversarial attacks with momentum,'' in \emph{Proceedings of the IEEE
  conference on computer vision and pattern recognition}, 2018, pp. 9185--9193.

\bibitem[Liu et~al.(2016)Liu, Chen, Liu, and Song]{liu2016delving}
Y.~Liu, X.~Chen, C.~Liu, and D.~Song, ``Delving into transferable adversarial
  examples and black-box attacks,'' \emph{arXiv preprint arXiv:1611.02770},
  2016.

\bibitem[Moosavi-Dezfooli et~al.(2017)Moosavi-Dezfooli, Fawzi, Fawzi, and
  Frossard]{moosavi2017universal}
S.-M. Moosavi-Dezfooli, A.~Fawzi, O.~Fawzi, and P.~Frossard, ``Universal
  adversarial perturbations,'' in \emph{Proceedings of the IEEE conference on
  computer vision and pattern recognition}, 2017, pp. 1765--1773.

\bibitem[Papernot et~al.(2017)Papernot, McDaniel, Goodfellow, Jha, Celik, and
  Swami]{papernot2017practical}
N.~Papernot, P.~McDaniel, I.~Goodfellow, S.~Jha, Z.~B. Celik, and A.~Swami,
  ``Practical black-box attacks against machine learning,'' in
  \emph{Proceedings of the 2017 ACM on asia conference on computer and
  communications security}, 2017, pp. 506--519.

\bibitem[Papernot et~al.(2016)Papernot, McDaniel, and
  Goodfellow]{papernot2016transferability}
N.~Papernot, P.~McDaniel, and I.~Goodfellow, ``Transferability in machine
  learning: from phenomena to black-box attacks using adversarial samples,''
  \emph{arXiv preprint arXiv:1605.07277}, 2016.

\bibitem[Szegedy et~al.(2013)Szegedy, Zaremba, Sutskever, Bruna, Erhan,
  Goodfellow, and Fergus]{szegedy2013intriguing}
C.~Szegedy, W.~Zaremba, I.~Sutskever, J.~Bruna, D.~Erhan, I.~Goodfellow, and
  R.~Fergus, ``Intriguing properties of neural networks,'' \emph{arXiv preprint
  arXiv:1312.6199}, 2013.

\bibitem[Tram{\`e}r et~al.(2017{\natexlab{a}})Tram{\`e}r, Papernot, Goodfellow,
  Boneh, and McDaniel]{tramer2017space}
F.~Tram{\`e}r, N.~Papernot, I.~Goodfellow, D.~Boneh, and P.~McDaniel, ``The
  space of transferable adversarial examples,'' \emph{arXiv preprint
  arXiv:1704.03453}, 2017.

\bibitem[Demontis et~al.(2019)Demontis, Melis, Pintor, Jagielski, Biggio,
  Oprea, Nita-Rotaru, and Roli]{demontis2019adversarial}
A.~Demontis, M.~Melis, M.~Pintor, M.~Jagielski, B.~Biggio, A.~Oprea,
  C.~Nita-Rotaru, and F.~Roli, ``Why do adversarial attacks transfer?
  explaining transferability of evasion and poisoning attacks,'' in
  \emph{Proceedings of the 28th USENIX conference on security symposium}, 2019,
  pp. 321--338.

\bibitem[Tram{\`e}r et~al.(2017{\natexlab{b}})Tram{\`e}r, Kurakin, Papernot,
  Goodfellow, Boneh, and McDaniel]{tramer2017ensemble}
F.~Tram{\`e}r, A.~Kurakin, N.~Papernot, I.~Goodfellow, D.~Boneh, and
  P.~McDaniel, ``Ensemble adversarial training: attacks and defenses,''
  \emph{arXiv preprint arXiv:1705.07204}, 2017.

\bibitem[Liu et~al.(2018)Liu, Simonyan, and Yang]{liu2018darts}
H.~Liu, K.~Simonyan, and Y.~Yang, ``Darts: differentiable architecture
  search,'' \emph{arXiv preprint arXiv:1806.09055}, 2018.

\bibitem[Devaguptapu et~al.(2020)Devaguptapu, Agarwal, Mittal, and
  Balasubramanian]{devaguptapu2020adversarial}
C.~Devaguptapu, D.~Agarwal, G.~Mittal, and V.~N. Balasubramanian, ``On
  adversarial robustness: a neural architecture search perspective,''
  \emph{arXiv preprint arXiv:2007.08428}, 2020.

\bibitem[Kotyan and Vargas(2019)]{kotyan2019evolving}
S.~Kotyan and D.~V. Vargas, ``Evolving robust neural architectures to defend
  from adversarial attacks,'' \emph{arXiv preprint arXiv:1906.11667}, 2019.

\bibitem[Shafahi et~al.(2019)Shafahi, Najibi, Ghiasi, Xu, Dickerson, Studer,
  Davis, Taylor, and Goldstein]{shafahi2019adversarial}
A.~Shafahi, M.~Najibi, A.~Ghiasi, Z.~Xu, J.~Dickerson, C.~Studer, L.~S. Davis,
  G.~Taylor, and T.~Goldstein, ``Adversarial training for free!'' \emph{arXiv
  preprint arXiv:1904.12843}, 2019.

\bibitem[Wong et~al.(2020)Wong, Rice, and Kolter]{wong2020fast}
E.~Wong, L.~Rice, and J.~Z. Kolter, ``Fast is better than free: revisiting
  adversarial training,'' \emph{arXiv preprint arXiv:2001.03994}, 2020.

\bibitem[Liu and Jin(2021)]{liu2021multi}
J.~Liu and Y.~Jin, ``Multi-objective search of robust neural architectures
  against multiple types of adversarial attacks,'' \emph{Neurocomputing}, vol.
  453, pp. 73--84, 2021.

\bibitem[Cai et~al.(2019)Cai, Gan, Wang, Zhang, and Han]{cai2019once}
H.~Cai, C.~Gan, T.~Wang, Z.~Zhang, and S.~Han, ``Once-for-all: train one
  network and specialize it for efficient deployment,'' \emph{arXiv preprint
  arXiv:1908.09791}, 2019.

\bibitem[Guo et~al.(2020)Guo, Yang, Xu, Liu, and Lin]{guo2020meets}
M.~Guo, Y.~Yang, R.~Xu, Z.~Liu, and D.~Lin, ``When nas meets robustness: in
  search of robust architectures against adversarial attacks,'' in
  \emph{Proceedings of the IEEE/CVF conference on computer vision and pattern
  recognition}, 2020, pp. 631--640.

\bibitem[Xie et~al.(2021)Xie, Wang, Yu, Zheng, and Jin]{xie2021tiny}
G.~Xie, J.~Wang, G.~Yu, F.~Zheng, and Y.~Jin, ``Tiny adversarial
  mulit-objective oneshot neural architecture search,'' \emph{arXiv preprint
  arXiv:2103.00363}, 2021.

\bibitem[Stanley and Miikkulainen(2002)]{stanley2002evolving}
K.~O. Stanley and R.~Miikkulainen, ``Evolving neural networks through
  augmenting topologies,'' \emph{Evolutionary computation}, vol.~10, no.~2, pp.
  99--127, 2002.

\bibitem[Sun et~al.(2020)Sun, Xue, Zhang, Yen, and Lv]{sun2020automatically}
Y.~Sun, B.~Xue, M.~Zhang, G.~G. Yen, and J.~Lv, ``Automatically designing cnn
  architectures using the genetic algorithm for image classification,''
  \emph{IEEE transactions on cybernetics}, vol.~50, no.~9, pp. 3840--3854,
  2020.

\bibitem[Jalwana et~al.(2020)Jalwana, Akhtar, Bennamoun, and
  Mian]{jalwana2020orthogonal}
M.~A. Jalwana, N.~Akhtar, M.~Bennamoun, and A.~Mian, ``Orthogonal deep models
  as defense against black-box attacks,'' \emph{IEEE Access}, vol.~8, pp.
  119\,744--119\,757, 2020.

\bibitem[Kariyappa and Qureshi(2019)]{kariyappa2019improving}
S.~Kariyappa and M.~K. Qureshi, ``Improving adversarial robustness of ensembles
  with diversity training,'' \emph{arXiv preprint arXiv:1901.09981}, 2019.

\bibitem[Kera and Hasegawa(2020)]{kera2020gradient}
H.~Kera and Y.~Hasegawa, ``Gradient boosts the approximate vanishing ideal,''
  in \emph{Proceedings of the 34th {AAAI} Conference on Artificial
  Intelligence}.\hskip 1em plus 0.5em minus 0.4em\relax AAAI Press, 2020, pp.
  4428--4425.

\bibitem[Kera(2021)]{kera2021border}
H.~Kera, ``Border basis computation with gradient-weighted norm,'' \emph{arXiv
  preprint arXiv:2101.00401}, 2021.

\bibitem[Kera and Hasegawa(2021)]{kera2021monomial}
H.~Kera and Y.~Hasegawa, ``Monomial-agnostic computation of vanishing ideals,''
  \emph{arXiv preprint arXiv:2101.00243}, 2021.

\bibitem[Krizhevsky et~al.(2009)Krizhevsky, Hinton,
  et~al.]{krizhevsky2009learning}
A.~Krizhevsky, G.~Hinton \emph{et~al.}, ``Learning multiple layers of features
  from tiny images,'' 2009.

\bibitem[LeCun et~al.(1998)LeCun, Bottou, Bengio, and
  Haffner]{lecun1998gradient}
Y.~LeCun, L.~Bottou, Y.~Bengio, and P.~Haffner, ``Gradient-based learning
  applied to document recognition,'' \emph{Proceedings of the IEEE}, vol.~86,
  no.~11, pp. 2278--2324, 1998.

\bibitem[Clanuwat et~al.(2018)Clanuwat, Bober-Irizar, Kitamoto, Lamb, Yamamoto,
  and Ha]{clanuwat2018deep}
T.~Clanuwat, M.~Bober-Irizar, A.~Kitamoto, A.~Lamb, K.~Yamamoto, and D.~Ha,
  ``Deep learning for classical japanese literature,'' \emph{arXiv preprint
  arXiv:1812.01718}, 2018.

\bibitem[He et~al.(2016)He, Zhang, Ren, and Sun]{he2016deep}
K.~He, X.~Zhang, S.~Ren, and J.~Sun, ``Deep residual learning for image
  recognition,'' in \emph{Proceedings of the IEEE conference on computer vision
  and pattern recognition}, 2016, pp. 770--778.

\bibitem[Simonyan and Zisserman(2014)]{simonyan2014very}
K.~Simonyan and A.~Zisserman, ``Very deep convolutional networks for
  large-scale image recognition,'' \emph{arXiv preprint arXiv:1409.1556}, 2014.

\bibitem[Huang et~al.(2017)Huang, Liu, Van Der~Maaten, and
  Weinberger]{huang2017densely}
G.~Huang, Z.~Liu, L.~Van Der~Maaten, and K.~Q. Weinberger, ``Densely connected
  convolutional networks,'' in \emph{Proceedings of the IEEE conference on
  computer vision and pattern recognition}, 2017, pp. 4700--4708.

\bibitem[Iandola et~al.(2016)Iandola, Han, Moskewicz, Ashraf, Dally, and
  Keutzer]{iandola2016squeezenet}
F.~N. Iandola, S.~Han, M.~W. Moskewicz, K.~Ashraf, W.~J. Dally, and K.~Keutzer,
  ``Squeezenet: alexnet-level accuracy with 50x fewer parameters and< 0.5 mb
  model size,'' \emph{arXiv preprint arXiv:1602.07360}, 2016.

\bibitem[Wu et~al.(2020)Wu, Wang, Xia, Bailey, and Ma]{wu2020skip}
D.~Wu, Y.~Wang, S.-T. Xia, J.~Bailey, and X.~Ma, ``Skip connections matter: on
  the transferability of adversarial examples generated with resnets,''
  \emph{arXiv preprint arXiv:2002.05990}, 2020.

\bibitem[Wu and Zhu(2020)]{wu2020towards}
L.~Wu and Z.~Zhu, ``Towards understanding and improving the transferability of
  adversarial examples in deep neural networks,'' in \emph{Asian conference on
  machine learning}.\hskip 1em plus 0.5em minus 0.4em\relax PMLR, 2020, pp.
  837--850.

\bibitem[Lyu et~al.(2015)Lyu, Huang, and Liang]{lyu2015unified}
C.~Lyu, K.~Huang, and H.-N. Liang, ``A unified gradient regularization family
  for adversarial examples,'' in \emph{2015 IEEE international conference on
  data mining}.\hskip 1em plus 0.5em minus 0.4em\relax IEEE, 2015, pp.
  301--309.

\bibitem[Ross and Doshi-Velez(2018)]{ross2018improving}
A.~S. Ross and F.~Doshi-Velez, ``Improving the adversarial robustness and
  interpretability of deep neural networks by regularizing their input
  gradients,'' in \emph{32nd AAAI conference on artificial intelligence}, 2018.

\bibitem[Zoph and Le(2016)]{zoph2016neural}
B.~Zoph and Q.~V. Le, ``Neural architecture search with reinforcement
  learning,'' \emph{arXiv preprint arXiv:1611.01578}, 2016.

\bibitem[Suganuma et~al.(2017)Suganuma, Shirakawa, and
  Nagao]{suganuma2017genetic}
M.~Suganuma, S.~Shirakawa, and T.~Nagao, ``A genetic programming approach to
  designing convolutional neural network architectures,'' in \emph{Proceedings
  of the genetic and evolutionary computation conference}, 2017, pp. 497--504.

\bibitem[Operiano et~al.(2020)Operiano, Iba, and
  Pora]{operiano2020neuroevolution}
K.~R.~G. Operiano, H.~Iba, and W.~Pora, ``Neuroevolution architecture backbone
  for x-ray object detection,'' in \emph{2020 IEEE symposium series on
  computational intelligence}.\hskip 1em plus 0.5em minus 0.4em\relax IEEE,
  2020, pp. 2296--2303.

\bibitem[Simonyan et~al.(2013)Simonyan, Vedaldi, and
  Zisserman]{simonyan2013deep}
K.~Simonyan, A.~Vedaldi, and A.~Zisserman, ``Deep inside convolutional
  networks: visualising image classification models and saliency maps,''
  \emph{arXiv preprint arXiv:1312.6034}, 2013.

\bibitem[Sundararajan et~al.(2017)Sundararajan, Taly, and
  Yan]{sundararajan2017axiomatic}
M.~Sundararajan, A.~Taly, and Q.~Yan, ``Axiomatic attribution for deep
  networks,'' in \emph{International conference on machine learning}.\hskip 1em
  plus 0.5em minus 0.4em\relax PMLR, 2017, pp. 3319--3328.

\bibitem[Madry et~al.(2017)Madry, Makelov, Schmidt, Tsipras, and
  Vladu]{madry2017towards}
A.~Madry, A.~Makelov, L.~Schmidt, D.~Tsipras, and A.~Vladu, ``Towards deep
  learning models resistant to adversarial attacks,'' \emph{arXiv preprint
  arXiv:1706.06083}, 2017.

\bibitem[Vargas and Murata(2016)]{vargas2016spectrum}
D.~V. Vargas and J.~Murata, ``Spectrum-diverse neuroevolution with unified
  neural models,'' \emph{IEEE transactions on neural networks and learning
  systems}, vol.~28, no.~8, pp. 1759--1773, 2016.

\bibitem[Agapie and Wright(2014)]{agapie2014theoretical}
A.~Agapie and A.~H. Wright, ``Theoretical analysis of steady state genetic
  algorithms,'' \emph{Applications of mathematics}, vol.~59, no.~5, pp.
  509--525, 2014.

\bibitem[Paszke et~al.(2019)Paszke, Gross, Massa, Lerer, Bradbury, Chanan,
  Killeen, Lin, Gimelshein, Antiga, Desmaison, Kopf, Yang, DeVito, Raison,
  Tejani, Chilamkurthy, Steiner, Fang, Bai, and Chintala]{paszke2019pytorch}
A.~Paszke, S.~Gross, F.~Massa, A.~Lerer, J.~Bradbury, G.~Chanan, T.~Killeen,
  Z.~Lin, N.~Gimelshein, L.~Antiga, A.~Desmaison, A.~Kopf, E.~Yang, Z.~DeVito,
  M.~Raison, A.~Tejani, S.~Chilamkurthy, B.~Steiner, L.~Fang, J.~Bai, and
  S.~Chintala, ``Pytorch: an imperative style, high-performance deep learning
  library,'' in \emph{Advances in neural information processing systems},
  H.~Wallach, H.~Larochelle, A.~Beygelzimer, F.~d\textquotesingle
  Alch\'{e}-Buc, E.~Fox, and R.~Garnett, Eds.\hskip 1em plus 0.5em minus
  0.4em\relax Curran Associates, Inc., 2019, pp. 8024--8035.

\bibitem[Shaikhina and Khovanova(2017)]{shaikhina2017handling}
T.~Shaikhina and N.~A. Khovanova, ``Handling limited datasets with neural
  networks in medical applications: a small-data approach,'' \emph{Artificial
  intelligence in medicine}, vol.~75, pp. 51--63, 2017.

\bibitem[Zhang and Ling(2018)]{zhang2018strategy}
Y.~Zhang and C.~Ling, ``A strategy to apply machine learning to small datasets
  in materials science,'' \emph{Npj computational materials}, vol.~4, no.~1,
  pp. 1--8, 2018.

\bibitem[Oh et~al.(2020)Oh, Park, and Ye]{oh2020deep}
Y.~Oh, S.~Park, and J.~C. Ye, ``Deep learning covid-19 features on cxr using
  limited training data sets,'' \emph{IEEE transactions on medical imaging},
  vol.~39, no.~8, pp. 2688--2700, 2020.

\bibitem[Wang et~al.(2020)Wang, Mo, Zhou, Xu, and Liu]{wang2020efficient}
D.~Wang, J.~Mo, G.~Zhou, L.~Xu, and Y.~Liu, ``An efficient mixture of deep and
  machine learning models for covid-19 diagnosis in chest x-ray images,''
  \emph{PloS one}, vol.~15, no.~11, p. e0242535, 2020.

\bibitem[Guo et~al.(2017)Guo, Rana, Cisse, and Van
  Der~Maaten]{guo2017countering}
C.~Guo, M.~Rana, M.~Cisse, and L.~Van Der~Maaten, ``Countering adversarial
  images using input transformations,'' \emph{arXiv preprint arXiv:1711.00117},
  2017.

\bibitem[Xie et~al.(2019)Xie, Wu, Maaten, Yuille, and He]{xie2019feature}
C.~Xie, Y.~Wu, L.~v.~d. Maaten, A.~L. Yuille, and K.~He, ``Feature denoising
  for improving adversarial robustness,'' in \emph{Proceedings of the IEEE/CVF
  conference on computer vision and pattern recognition}, 2019, pp. 501--509.

\bibitem[Xiao et~al.(2017)Xiao, Rasul, and Vollgraf]{xiao2017fashion}
H.~Xiao, K.~Rasul, and R.~Vollgraf, ``Fashion-mnist: a novel image dataset for
  benchmarking machine learning algorithms,'' \emph{arXiv preprint
  arXiv:1708.07747}, 2017.

\end{thebibliography}


\end{document}